\documentclass[11pt,a4paper]{article}
\usepackage[hyperref]{acl2020}
\usepackage{times}
\usepackage{latexsym}

\usepackage{caption}
\usepackage{subcaption}
\usepackage{graphicx}
\usepackage{booktabs}

\usepackage{amsmath}
\usepackage{adjustbox}

\usepackage{multirow}
\usepackage{longtable}

\usepackage[utf8]{inputenc}

\usepackage{microtype}

\usepackage{array}
\usepackage{xspace}

\usepackage{xcolor,colortbl}
\usepackage{tikz}
\usepackage{collcell}
\usepackage{xstring}

\usepackage{scalerel}

\usepackage{float}
\newfloat{footnote1}{hb}

\aclfinalcopy % Uncomment this line for the final submission
\title{GLUECoS : An Evaluation Benchmark for Code-Switched NLP}

\author{Simran Khanuja\textsuperscript{1} \quad Sandipan Dandapat\textsuperscript{2} \quad \textbf{Anirudh Srinivasan\textsuperscript{1}}\\ \quad \textbf{Sunayana Sitaram\textsuperscript{1}} \quad \textbf{Monojit Choudhury\textsuperscript{1}}\\
\textsuperscript{1} Microsoft Research, Bangalore, India \\
\textsuperscript{2} Microsoft R\&D, Hyderabad, India \\
\{t-sikha, sadandap, t-ansrin, sunayana.sitaram, monojitc\}@microsoft.com \\
}

\newcommand{\vbl}{{\vrule width 1.1pt}}
\newcommand{\hbl}{\noalign{
\hrule height 1.1pt
}}
\newcommand{\pad}{1.3}

\date{}

\begin{document}
\maketitle
\begin{abstract}
Code-switching is the use of more than one language in the same conversation or utterance. Recently, multilingual contextual embedding models, trained on multiple monolingual corpora, have shown promising results on cross-lingual and multilingual tasks. We present an evaluation benchmark, GLUECoS, for code-switched languages, that spans several NLP tasks in English-Hindi and English-Spanish. Specifically, our evaluation benchmark includes Language Identification from text, POS tagging, Named Entity Recognition, Sentiment Analysis, Question Answering and a new task for code-switching, Natural Language Inference. We present results on all these tasks using cross-lingual word embedding models and multilingual models. In addition, we fine-tune multilingual models on artificially generated code-switched data. Although multilingual models perform significantly better than cross-lingual models, our results show that in most tasks, across both language pairs, multilingual models fine-tuned on code-switched data perform best, showing that multilingual models can be further optimized for code-switching tasks.
\end{abstract}

\section{Introduction}
Code-switching, or code-mixing, is the use of more than one language in the same utterance or conversation and is prevalent in multilingual societies all over the world. It is a spoken phenomenon and is found most often in informal chat and social media on the Internet. Processing, understanding, and generating code-mixed text and speech has become an important area of research.

Recently, contextual word embedding models trained on a large amount of text data have shown state-of-the-art results in a variety of NLP tasks. Models such as BERT \cite{devlin2018bert} and its multilingual version, mBERT, rely on large amounts of unlabeled monolingual text data to build monolingual and multilingual models that can be used for downstream tasks involving limited labelled data. \cite{wang2018glue} propose a Generalized Language Evaluation Benchmark (GLUE) to evaluate embedding models on a wide variety of language understanding tasks. This benchmark has spurred research in monolingual transfer learning settings.

Data and annotated resources are scarce for code-switched languages, even if one or both languages being mixed are high resource. Due to this, there is a lack of standardized datasets in code-switched languages other than those used in shared tasks in a few language pairs. Although models using synthetic code-switched data and cross-lingual embedding techniques have been proposed for code-switching \cite{pratapa2018language}, there has not been a comprehensive evaluation of embedding models across different types of tasks. Furthermore, there have been claims that multilingual models such as mBERT are competent in zero-shot cross lingual transfer and code-switched settings. Though comprehensively validated by \cite{pires2019multilingual} in the case of zero-shot transfer, the probing in code-switched settings was limited to one dataset of one task, namely POS Tagging.

To address all these issues and inspired by the GLUE \cite{wang2018glue} benchmark, we propose GLUECoS, a language understanding evaluation framework for \textbf{Co}de-\textbf{S}witched NLP. We include five tasks from previously conducted evaluations and shared tasks, and propose a sixth, \textit{Natural Language Inference} task for code-switching, using a new dataset\footnote{we use a subset of the original corpus as available to us at the time of experimentation}\cite{khanuja2020new}. We include tasks varying in complexity ranging from word-level tasks [\textit{Language Identification (LID); Named Entity Recognition (NER)}], syntactic tasks [\textit{POS tagging}], semantic tasks [\textit{Sentiment Analysis; Question Answering}] and finally a \textit{Natural Language Inference} task. Where available, we include multiple datasets for each task in English-Spanish and English-Hindi. We choose these language pairs, not only due to the relative abundance of publicly available datasets, but also because they represent variations in types of code-switching, language families, and scripts between the languages being mixed.
% We choose these language pairs, not only due to the availability of public datasets in these languages, but also because they represent differences in types of code-switching, language families and script differences between the languages being mixed. 
We test various cross-lingual and multilingual models on all of these tasks. In addition, we also test models trained with synthetic code-switched data. Lastly, we fine-tune the best performing multilingual model with synthetic code-switched data and show that in most cases, its performance exceeds the multilingual model, highlighting that multilingual models can be further optimized for code-switched settings.

The main contributions of our work are as follows:
\begin{itemize}
\item{We point out the lack of standardized datasets for code-switching and propose an evaluation benchmark GLUECoS, which can be used to test models on various NLP tasks in English-Hindi and English-Spanish.}

\item{In creating the benchmark, we highlight the tasks that are missing from code-switched NLP and propose a new task, \textit{Natural Language Inference}, for code-switched data.}

\item{We evaluate cross-lingual and pre-trained multilingual embeddings on all these tasks, and observe that pre-trained multilingual embeddings significantly outperform cross-lingual embeddings. This highlights the competence of generalized language models over cross lingual word embeddings.}

\item{We fine-tune pre-trained multilingual models on linguistically motivated synthetic code-switched data, and observe that they perform better in most cases, highlighting that these models can be further optimized for code-switched settings.}
% 1) point out lack of standardized datasets and a call for the community to agree upon these for the same.\\
% 2) point out the incompetence of clwe for processing code-mixed which is expected given that they do not optimize for code-mixed evaluation and we should work towards the same.\\
% 3) highlight the power of mBERT and modified mBERT and motivation to build a customized generalized language model for cm which is pre-trained on large amounts of monoligual corpora of langauges involved + gcm +rcm just like adithya's paper's LM, which we believe would be ideal to process cm data given the promising direction offered by modified mbert

% 4) highlight difference in en-hi and en-es and say how different language families code-mixed may need more attention/more data/ different ways of processing etc.\\
% 5) highlight how some tasks for cm are missing attention from the community like NLI \\
\end{itemize}
% Why it is harder to do some tasks than others (syntactic vs semantic)

% How much data is there - low resource angle

% A little about GLUE and why we decided to do a CM version

% NLI dataset creation

% mBERT, CL word embeddings, ...

The rest of the paper is organized as follows. We relate our work to prior work to situate our contributions. We introduce the tasks and datasets used for GLUECoS motivating the choices we make. We describe the experimental setup, with details of the models used for baseline evaluations. We present the results of testing all the models on the benchmark and analyze the results. We conclude with a direction for future work and highlight our main findings.

\section{Relation to prior work}

The idea of a generalized benchmark for code-switching is inspired by GLUE \cite{wang2018glue}, which has spurred research in \textit{Natural Language Understanding} in English, to an extent that a set of harder tasks have been curated in a follow-up benchmark, SuperGLUE \cite{wang2019superglue} once models beat the human baseline for GLUE. The motivation behind GLUE is to evaluate models in a multi-task learning framework across several tasks, so that tasks with less training data can benefit from others. Although our current work does not include models evaluated in a multi-task setting, we plan to implement this in subsequent versions of the benchmark.

There have been shared tasks conducted in the past as part of code-switching workshops co-located with notable NLP conferences. The first and second workshops on \textit{Computational Approaches to Code Switching} \cite{diab2014proceedings, diab2016proceedings}  conducted a shared task on \textit{Language Identification} for several language pairs \cite{solorio2014overview, molina2016overview}. The third workshop \cite{aguilar2018proceedings} included a shared task on \textit{Named Entity Recognition} for the English-Spanish and Modern Standard Arabic-Egyptian Arabic language pairs\cite{aguilar2019named}. 

The \textit{Forum for Information Retrieval Evaluation} (FIRE) aims to meet new challenges in multilingual information access and has conducted several shared tasks on code-switching. These include tasks on \textit{transliterated search}, \cite{roy2013overview, choudhury2014overview} \textit{code-mixed entity extraction} \cite{rao2016cmee} and \textit{mixed script information retrieval} \cite{sequiera2015overview, banerjee2016overview}. Other notable shared tasks include the \textit{Tool Contest on POS Tagging for Code-Mixed Indian Social Media} at ICON 2016 \cite{jamatia2016collecting}, \textit{Sentiment Analysis for Indian Languages (Code-Mixed)} at ICON 2017 \cite{patra2018sentiment} and the \textit{Code-Mixed Question Answering Challenge} \cite{chandu2018code}.

Each of the shared tasks mentioned above attracted several participants and have led to follow up research in these problems. However, all tasks have focused on a single NLP problem and so far, there has not been an evaluation of models across several code-switched NLP tasks. Our objective with proposing GLUECoS is to address this gap, and determine which models best generalize across different tasks, languages and datasets.

\section{Tasks and Datasets}

Some NLP tasks are inherently more complex than others - for example, a \textit{Question Answering} task that needs to understand both the meaning of the question and answer, is harder to solve by a machine than a word-level \textit{Language Identification} task, in which a dictionary lookup can give reasonable results. Some datasets and domains may contain very little code-switching, while others may contain more frequent and complex code-switching. Similar languages, when code-switched, may maintain the word order of both languages, while other language pairs that are very different may take on the word order of one of the languages. With these in mind, our choice of tasks and datasets for GLUECoS are based on the following principles :

% We wanted a variety of tasks, ranging from easier ones, on which the research community has already achieved very high accuracies, to more difficult ones on which very few attempts have been made. Secondly, we wanted to evaluate models on a variety of languages from different language families, but wanted to include a number of tasks in the same language pair to enable comparisons across tasks. This led us to choose English-Hindi and English-Spanish, for which we found datasets for almost all the tasks in our benchmark. English and Spanish are written in the Roman script, while English-Hindi datasets can contain Hindi words written either in the original Devanagari script, or in the Roman script. We included multiple datasets from each language pair when available for a task, so that results can be compared across datasets. 
\begin{itemize}
    \item We choose a variety of tasks, ranging from simpler ones, on which the research community has already achieved high accuracies, to relatively more complex, on which very few attempts have been made.
    \item We desire to evaluate models on language-pairs from different language families, and on a varied number of tasks, to enable detailed analysis and comparison. This led us to choose \textit{English-Hindi} and \textit{English-Spanish}, as we found researched upon datasets for almost all tasks in our benchmark for these language pairs.
    \item English and Spanish are written in the Roman script, while \textit{English-Hindi} datasets can contain Hindi words written either in the original Devanagari script, or in the Roman script, thus adding script variance as an additional parameter to analyse upon.
    \item  We include multiple datasets from each language pair where available, so that results can be compared across datasets for the same task.
\end{itemize}
Due to the lack of standardized datasets, we choose to create our own train-test-validation splits for some tasks. Also, we use an off-the-shelf transliterator and language detector, where necessary, details of which can be found in Appendix \ref{app:dataset}. Table \ref{Corpus-stats} shows all the datasets that we use, with their statistics, while Table \ref{Data-stats} shows the code-switching statistics of the data in terms of standardized metrics for code-switching \cite{gamback2014measuring, guzman2017metrics}. Briefly, the code-mixing metrics include :
\begin{itemize}
    \item \textit{Code-Mixing Index} (CMI) : The fraction of language dependent tokens not belonging to the matrix language in the utterance.
    \item \textit{Average switch-points} (SP Avg) : The average number of intra-sentential language switch-points in the corpus. 
    \item \textit{Multilingual Index} (M-index) : A word-count-based measure quantifying the inequality of distribution of language tags in a corpus of at least two languages.
    \item \textit{Probability of Switching} (I-index) :The proportion of the number of switchpoints in the corpus, relative to the number of language-dependent tokens.
    \item \textit{Burstiness} : The quantification of whether switching occurs in bursts (randomly similar to a Poisson process), or has a more periodic character.
    \item \textit{Language Entropy} (LE) : The bits of information needed to describe the distribution of language tags.
    \item \textit{Span Entropy} (SE) : The bits of information needed to describe the distribution of language spans.
\end{itemize}

% There exists a lack of standardized datasets here because of which we had to choose our own training data and create custom splits for a few tasks wherein the data wasn't made public. As far as possible, we have tried to choose datasets involved in shared tasks since they would be better curated for a task, but we couldn't obtain such datasets for all tasks (eg - en-hi NER) and many shared tasks didn't release test labels or training data (fire 13 en-es ner). These inconsistencies made us resort to our own choices of training data which we have described above.
In cases where the datasets have been a part of shared tasks, we report the highest scores obtained in each task as the \textit{State Of The Art} (SOTA) for the dataset. However, note that we report this to situate our results in context of the same, and these cannot be directly compared, since each task's SOTA is obtained by varied training architecture, suited to perform well in one particular task alone.

\begin{table*}
\centering
\small
{\renewcommand{\arraystretch}{\pad}
\begin{tabular}{! \vbl >{\columncolor[HTML]{EFEFEF}}p{2.5cm}|p{1.7 cm}|p{1.6cm}|p{1.6cm}|p{1.6cm}! \vbl}
 \hbl
 \multicolumn{5}{! \vbl c! \vbl}{\cellcolor[HTML]{C0C0C0}\textbf{English-Hindi}} \\
 \hline
 \textbf{Corpus} & \cellcolor[HTML]{EFEFEF}\textbf{Sent (Train)}&\cellcolor[HTML]{EFEFEF}\textbf{Sent (Dev)}&\cellcolor[HTML]{EFEFEF}\textbf{Sent (Test)}&\cellcolor[HTML]{EFEFEF}\textbf{Sent (All)}\\
 \hline
 Fire LID (D)  & 2631  & 500  & 406  & 3537  \\ \hline
 UD POS (D)  & 1384  & 215  & 215  & 1814  \\ \hline
 FG POS (R) & 2104  & 263  & 264  & 2631 \\ \hline
 IIITH NER (R)  & 2467 & 308 & 309 & 3084 \\ \hline
 SAIL Sentiment (R) & 10080 & 1260 & 1261 & 12601  \\ \hline
 %Parsing  & - & - & - & - & - & - & - & - & -\\
 QA (R)  & 250  & -  & 63  & 313  \\ \hline
 NLI (R) & 1040 & 130 & 130 & 1300  \\
 \hline
  \multicolumn{5}{! \vbl c! \vbl}{\textbf{\cellcolor[HTML]{C0C0C0} English-Spanish}} \\
 \hline
 \textbf{Corpus} & \cellcolor[HTML]{EFEFEF}\textbf{Sent (Train)}&\cellcolor[HTML]{EFEFEF}\textbf{Sent (Dev)}&\cellcolor[HTML]{EFEFEF}\textbf{Sent (Test)}&\cellcolor[HTML]{EFEFEF}\textbf{Sent (All)}\\
 \hline
 EMNLP 2014  & 10259 & 1140 & 3014 & 14413  \\ \hline
 Bangor POS  & 2192 & 274 & 274 & 2758 \\ \hline
 CALCS NER  & 27366 & 3420 &  3421& 34208 \\ \hline
 Sentiment  & 1681 & 211  & 211  & 2103   \\ 
 %Parsing  & - & - & - & - & - & - & - & - \\
 \hbl
\end{tabular}}
\caption{\label{Corpus-stats} Corpus Statistics. (R) and (D) indicates Hindi written in Roman and Devanagari script, respectively  }
\end{table*}

\begin{table*}[h]
\centering
\small
{\renewcommand{\arraystretch}{\pad}
\begin{tabular}{! \vbl >{\columncolor[HTML]{EFEFEF}}p{2.2cm}|p{1.2cm}|p{1.2cm}|p{1.2cm}|p{1.2cm}|p{1.35cm}|p{1.2cm}|p{1.2cm}! \vbl}
 \hbl
 \multicolumn{8}{! \vbl c! \vbl}{\cellcolor[HTML]{C0C0C0}\textbf{English-Hindi}} \\
 \hline
 \textbf{Corpus} & \cellcolor[HTML]{EFEFEF}\textbf{CMI} & \cellcolor[HTML]{EFEFEF}\textbf{SP Avg} & \cellcolor[HTML]{EFEFEF}\textbf{M-index} & \cellcolor[HTML]{EFEFEF}\textbf{I-index} & \cellcolor[HTML]{EFEFEF}\textbf{Burstiness}  & \cellcolor[HTML]{EFEFEF}\textbf{LE} & \cellcolor[HTML]{EFEFEF}\textbf{SE}\\
 \hline
 Fire LID  & 78.26 & 4.47 & 0.39 & 0.33 & -0.42 & 0.86 & 1.02 \\ \hline
 UD POS  & 136 & 4.98 & 0.46 & 0.39 & -0.25 & 1.35 & 1.47 \\ \hline
 FG POS  & 68 & 5.5 & 0.4 & 0.34 & -0.43 & 0.87 & 1.05 \\ \hline
 IIITH NER  & 133 & 11.39 & 0.64 & 0.53 & -0.26 & 1.28 & 1.36  \\ \hline
 SAIL Sentiment& 72.8 & 5.07 & 0.02 & 0.32 & -0.32 & 0.87 & 1.17  \\ \hline
 %Parsing  & - & - & - & - & - & - & - & - \\
 QA  & 142.28 & 3.96 & 0.81 & 0.5 & -0.4 & 0.89 & 1.09  \\ \hline
 NLI  & 149.95 & 66.74 & 0.44 & 0.63 & -0.2 & 1.53 & 1.39  \\
 \hline
  \multicolumn{8}{! \vbl c! \vbl}{\cellcolor[HTML]{C0C0C0}\textbf{English-Spanish}} \\
 \hline
 \textbf{Corpus} & \cellcolor[HTML]{EFEFEF}\textbf{CMI} & \cellcolor[HTML]{EFEFEF}\textbf{SP Avg} & \cellcolor[HTML]{EFEFEF}\textbf{M-index} & \cellcolor[HTML]{EFEFEF}\textbf{I-index} & \cellcolor[HTML]{EFEFEF}\textbf{Burstiness} & \cellcolor[HTML]{EFEFEF}\textbf{LE} & \cellcolor[HTML]{EFEFEF}\textbf{SE}\\
 \hline
 EMNLP 2014 & 33.46 & 2.86 & 0.33 & 0.29 & -0.34 & 0.79 & 1.1  \\ \hline
 Bangor POS  & 123.06 & 1.67 & 0.32 & 0.27 & -0.35 & 0.82 & 1.06 \\ \hline
 CALCS NER  & 94.52 & 3.17 & 0.004 & 0.31 & -0.42 & 0.75 & 1.02  \\ \hline
Sentiment  & 110.56 & 4.13 & 0.15 & 0.27 & -0.21 & 0.79 & 1.42  \\
 %Parsing  & - & - & - & - & - & - & - & - \\
 \hbl
\end{tabular}}
\caption{\label{Data-stats} Code-switching Statistics }
\end{table*}

% \subsection{POS}
% *Brief description of POS. 2 tables, one describing the corpus' chosen for POS and another for the results on several baselines*
% \begin{tabular}{ |p{1.6cm}||p{1.4cm}|p{1.4cm}|p{1.4cm}|p{1.4cm}|p{1.5cm}|p{1.4cm}|p{1.4cm}|p{1.4cm}| }
%  \hline
%  Corpus & Language Pair & Language Tag & Sentences(Train/Dev) & Tokens(Train/Dev) & Tokens(Test) & Tokens(Test) & Number of types\\
%  \hline
%  Bangor & - & - & - & - & - & - & - & - \\
%  Amitava-ICON  & - & - & - & - & - & - & - & - \\
%  \hline
% \end{tabular}
% Given below is a brief description of each task and the datasets chosen for the same :
\subsection{Language Identification (LID)}

\textit{Language Identification} is the task of obtaining word-level language labels for code-switched sentences. For \textit{English-Hindi} we choose the FIRE 2013 (FIRE LID) dataset originally created for the transliterated search subtask \cite{roy2013overview}. The test and development sets provided contain word-level language tagged sentences. For training we use a POS tagging dataset \cite{jamatia2016collecting} which also contains language labels.
\vskip 1mm
% We choose this dataset because the Hindi words were in both Roman and Devanagari giving us the freedom to experiment with models and embeddings trained on either. We report our results in this paper on the Devanagari version of the same since mBERT is trained on Devanagari. 

For \textit{English-Spanish} we choose the dataset in \cite{solorio2014overview}, provided as part of the LID shared task at EMNLP 2014. We report the highest score obtained for \textit{SPA-EN} \cite{solorio2014overview} as the SOTA for this task.

\subsection{Part of Speech (POS) tagging}

\textit{POS tagging} includes labelling at the word level,  grammatical part of speech tags such as noun, verb, adjective, pronoun, prepositions etc. For \textit{English-Hindi}, we use two datasets. The first is the code-switched Universal Dependency parsing dataset provided by \cite{bhat2018universal} (UD POS). This corpus contains a {\it transliterated} version, where Hindi is in the Roman script, and also a {\it corrected} version in which Hindi has been manually converted back to Devanagari. We report the highest score obtained by \cite{bhat2018universal} as the SOTA for this task.
% We conduct our experiments on the {\it corrected} version of the same.\\

The second \textit{English-Hindi} dataset we use was part of the \textit{ICON 2016 Tool Contest on POS Tagging for Code-Mixed Indian Social Media Text} \cite{jamatia2016collecting} (FG POS). We report the highest score obtained by
\cite{keyA}- (report communicated directly by authors) as the SOTA for this task.
% \footnote{https://docs.microsoft.com/en-us/azure/cognitive-services/translator/reference/v3-0-transliterate} to transliterate Hindi into Devanagari.

For \textit{English-Spanish}, of the two corpora utilised in \cite{alghamdi2016part}, we choose the Bangor Miami corpus (Bangor POS) owing to the larger size of the corpus. We report the highest score obtained by \cite{alghamdi2016part} as the SOTA for this task.
\subsection{Named Entity Recognition (NER)}
NER involves recognizing named entities such as {\it person, location, organization etc.} in a segment of text. For \textit{English-Hindi} we use the Twitter NER corpus provided by \cite{singh2018named} (IIITH NER). We report the highest score obtained by \cite{singh2018named} as the SOTA for this task.

% We report the F1 scores obtained on the transliterated corpus.\\

For \textit{English-Spanish}, we use the Twitter NER corpus provided as part of the CALCS 2018 shared task on NER for code-switched data \cite{aguilar2019named} (CALCS NER). We report the highest score obtained by \cite{winata2019hierarchical} as the SOTA for this task.

\vskip 1mm
% Note that we create a custom split of the training data available to us while the shared task numbers are on the test data.

% removing this as it shows we are from MS \footnote{https://docs.microsoft.com/en-us/azure/cognitive-services/translator/reference/v3-0-detect}

\subsection{Sentiment Analysis}

\textit{Sentiment analysis} is a sentence classification task wherein each sentence is labeled to be expressing a {\it positive, negative} or \textit{neutral} sentiment.

For \textit{English-Hindi} we choose the sentiment annotated social media corpus used in the \textit{ICON 2017 shared task;  Sentiment Analysis for Indian Languages (SAIL)} \cite{patra2018sentiment}. This corpus is originally language tagged at the word level with Hindi in the Roman script. We report the highest score obtained for \textit{HI-EN} \cite{patra2018sentiment} as the SOTA for this task.

For \textit{English-Spanish} we choose the sentiment annotated Twitter dataset provided by \cite{vilares2016cs} which we split into an 8:1:1 train:test:validation split ensuring sentiment distribution. \cite{vilares2016cs} report an average F1 score of 58.9 on the same dataset, while \cite{pratapa2018word} report an F1 of 64.6 on the same, which we report as the SOTA for this dataset. We are not aware of future work done on this dataset.
% This corpus is not language tagged, hence to evaluate the code-switching metrics of this corpus, we obtain word level language tags using an off-the-shelf language detector.  

\subsection{Question Answering (QA)}

\textit{Question Answering} is the task of answering a question based on the given context or world knowledge. We choose the dataset provided by~\cite{chandu2018code} which contains two types of questions for En-Hi, one with context (185 article based questions) and one containing image based questions (774 questions). For the image based questions we use the \textit{DrQA - Document Retriever module}\footnote{https://github.com/facebookresearch/DrQA} to extract the most relevant context from Wikipedia. Since it is a code-switched dataset, context could not be extracted for all questions. We obtain a final dataset having 313 (question-answer-context) triples.

% We then make use of the transformer library\footnote{https://github.com/huggingface/transformers} and the BiDAF architecture\footnote{https://github.com/ElizaLo/Question-Answering-based-on-SQuAD} for evaluation on BERT and clwe respectively.

\subsection{Natural Language Inference (NLI)}

\textit{Natural Language Inference} is the task of inferring a positive (entailed) or negative (contradicted) relationship between a premise and hypothesis. While most NLI datasets contain sentences or images as premises, the code-switched NLI dataset we use contains conversations as premises, making it a conversational NLI task \cite{khanuja2020new}. Since this is a new dataset, we report our number as the SOTA for this task.

% % rename section, describe how we created the NLI dataset
% % First paragraph - importance of NLI and motivation
% An important component of evaluating most natural language understanding systems is how well they perform on inference tasks.... 
% %Second paragraph - description of construction of dataset

\section{Experimental Setup}

%  ***Add a little about cross lingual tasks' nature and stuff***.

We use standard architectures for solving each of the tasks mentioned above (Refer to Appendix \ref{app:train}). We experiment with several existing cross lingual word embeddings that have been shown to perform well on cross lingual tasks. We also experiment with the Multilingual BERT (mBERT) model released by \cite{devlin2018bert}. In a survey on cross lingual word embeddings, \cite{ruder2017survey} establish that various embedding methods optimize for similar objectives given that the supervision data involved in training them is similar. Based on this, we choose the following representative embedding methods that vary in the amount of supervision involved in training them.

\subsection{MUSE Embeddings}

We use the MUSE library\footnote{https://github.com/facebookresearch/MUSE} to train both supervised and unsupervised word embeddings. The unsupervised word embeddings are learnt without any parallel data or anchor point. It learns a mapping from the source to the target space using adversarial training and (iterative) Procrustes refinement~\cite{conneau2017word}. The supervised method leverages a bilingual dictionary (or identical character strings as anchor points), to learn a mapping from the source to the target space using (iterative) Procrustes alignment.

% ***include what tasks muse performs well in?***
\subsection{BiCVM Embeddings}
This method, proposed by \cite{hermann-blunsom-2014-multilingual}, leverages parallel data, based on the assumption that parallel sentences are equivalent in meaning and subsequently have similar sentence representations. 
% They optimize for the sentence representations of aligned sentences to be closer than for randomly chosen negative samples, along with the monolingual objectives. 
We use the BiCVM toolkit\footnote{https://github.com/karlmoritz/bicvm/} to learn these embeddings. The parallel corpus we use for \textit{English-Spanish} consists of 4.5M parallel sentences from Twitter. For \textit{English-Hindi}, we make use of an internal parallel corpus consisting of roughly 5M parallel sentences. 

% ***include what bicvm performs well in? ***  
\subsection{BiSkip Embeddings}
This method makes use of parallel corpora as well as word alignments to learn cross-lingual embeddings. \cite{luong2015bilingual} adapt the skip-gram objective originally proposed by \cite{mikolov2013distributed} to a bilingual setting wherein a model learns to predict words cross-lingually along with the monolingual objectives. We make use of the \textit{fastalign toolkit}\footnote{https://github.com/clab/fast\_align} to learn word alignments given parallel corpora and use the \textit{BiVec toolkit}\footnote{https://github.com/lmthang/bivec} to learn the final BiSkip embeddings given the parallel corpora and the word alignments. The parallel corpora utilised to learn these are the same as those used to learn the BiCVM embeddings. 

% ***include what biskip does well in?***
\subsection{Synthetic Data (GCM) Embeddings}
We also experiment with skip-gram embeddings learnt from synthetically generated code-mixed data as proposed by \cite{pratapa2018word}. We make use of the \textit{fasttext library}\footnote{https://fasttext.cc/} to learn the skip-gram embeddings. For \textit{English-Spanish}, we obtain data from \cite{pratapa2018language} which consists of 8M synthetic code-switched sentences. For \textit{English-Hindi}, we generate synthetic data from the IITB parallel corpus.\footnote{http://www.cfilt.iitb.ac.in/iitb\_parallel/} We sample from the generated sentences obtained using Switch Point Fraction (SPF), as described in \cite{pratapa2018language}, to obtain a GCM corpus of roughly 10M sentences. 

\subsection{mBERT}
Multilingual BERT is pre-trained on monolingual corpora of 104 languages and has been shown to perform well on zero shot cross-lingual model transfer and code-switched POS tagging \cite{pires2019multilingual}. Specifically, we use the \textit{bert-base-multilingual-cased} model for our experiments.

% We kept a consistent learning rate of 5e-5, adam optimizer, and a batch size of 32/64 for word level/sentence level tasks; parameters proposed in \cite{devlin2018bert}
\subsection{Modified mBERT}
\cite{sun2019fine} show that fine-tuning BERT with in-domain data on language modeling improves performance on downstream tasks. On similar lines, we fine-tune the mBERT model with synthetically generated code-switched data (gCM) and a small amount of real code-switched data (rCM), on the masked language modeling objective. The training curriculum we use in fine-tuning this model is similar to as proposed by \cite{pratapa2018language}, which has been shown to improve language modeling perplexity.
% We first fine-tune it on synthetic code-mixed data for 20 epochs followed by fine-tuning for 20 epochs on real code-mixed data. 
Although we train on real code-mixed data, it accounts for a small fraction (less than 5\%) of the total code-mixed data used. Refer to Appendix \ref{app:bert} for training details.

% The data used for this is the same as the one used to learn skipgram for gcm ***ask anirudh for more details*** 

\begin{table*}[!htp]
\centering
\small
{\renewcommand{\arraystretch}{\pad}
\begin{tabular}{! \vbl l|c|c|c|c|l! \vbl}
\hbl 
 \cellcolor[HTML]{C0C0C0}\textbf{Data} & \cellcolor[HTML]{EFEFEF}\textbf{Baseline} & \cellcolor[HTML]{EFEFEF}\textbf{Unsup. MUSE} & \cellcolor[HTML]{EFEFEF}\textbf{Sup. MUSE} & \cellcolor[HTML]{EFEFEF}\textbf{BiSkip} &\cellcolor[HTML]{EFEFEF}\textbf{SOTA} \\
 \hline
 \cellcolor[HTML]{EFEFEF} & 93.21 & 94.53 & 94.92 & 93.98 & \\
  \cline{2-5}
 \cellcolor[HTML]{EFEFEF}& \cellcolor[HTML]{EFEFEF}\textbf{BiCVM} & \cellcolor[HTML]{EFEFEF}\textbf{GCM} & \cellcolor[HTML]{EFEFEF}\textbf{mBERT} & \cellcolor[HTML]{EFEFEF}\textbf{Mod. mBERT} &  \\
 \cline{2-5}
 \cellcolor[HTML]{EFEFEF} \multirow{-3}{*}{FIRE En-Hi} & 95.24 & 93.64 & 95.87 & \textbf{96.6} & \multirow{-3}{*}{N/A\footnotemark}\\
%  \footnote{Since the original task was language tagging and transliteration of Hindi words in the roman script while we report LID results for Hindi in Devanagari we report this as N/A. An accuracy of 99.0 was obtained on the original subtask\cite{roy2013overview}}
 \hline
  \cellcolor[HTML]{EFEFEF} & \cellcolor[HTML]{EFEFEF}\textbf{Baseline} & \cellcolor[HTML]{EFEFEF}\textbf{Unsup. MUSE} & \cellcolor[HTML]{EFEFEF}\textbf{Sup. MUSE} & \cellcolor[HTML]{EFEFEF}\textbf{BiSkip}  &\cellcolor[HTML]{EFEFEF}\textbf{SOTA}\\
 \cline{2-6}
  \cellcolor[HTML]{EFEFEF} & 92.95 & 92.86 & 93.39 & 92.79 & \\
  \cline{2-5}
 \cellcolor[HTML]{EFEFEF} & \cellcolor[HTML]{EFEFEF}\textbf{BiCVM} & \cellcolor[HTML]{EFEFEF}\textbf{GCM} & \cellcolor[HTML]{EFEFEF}\textbf{mBERT} & \cellcolor[HTML]{EFEFEF}\textbf{Mod. mBERT}& \\
 \cline{2-5}
  \cellcolor[HTML]{EFEFEF} \multirow{-4}{*}{EMNLP En-Es} & 91.47 & 92.42 & 95.97 & \textbf{96.24} & \multirow{-3}{*}{94.0}  \\
\hbl
\end{tabular}}
\caption{\label{LID-results} LID results (F1) }
\end{table*}

\footnotetext{The original task was language tagging and transliteration of Hindi words in the Roman script, while we report LID results for Hindi in Devanagari. An accuracy of 99.0 was obtained on the original subtask\cite{roy2013overview}}

\begin{table*}
\centering
\small
{\renewcommand{\arraystretch}{\pad}
\begin{tabular}{! \vbl l|c|c|c|c|l! \vbl}
\hbl 
 \cellcolor[HTML]{C0C0C0}\textbf{Data} & \cellcolor[HTML]{EFEFEF}\textbf{Baseline} & \cellcolor[HTML]{EFEFEF}\textbf{Unsup. MUSE} & \cellcolor[HTML]{EFEFEF}\textbf{Sup. MUSE} & \cellcolor[HTML]{EFEFEF}\textbf{BiSkip} &\cellcolor[HTML]{EFEFEF}\textbf{SOTA} \\
 \hline
 \cellcolor[HTML]{EFEFEF} & 77.49 & 78.06 & 77.88 & 77.43 & \\
  \cline{2-5}
 \cellcolor[HTML]{EFEFEF}& \cellcolor[HTML]{EFEFEF}\textbf{BiCVM} & \cellcolor[HTML]{EFEFEF}\textbf{GCM} & \cellcolor[HTML]{EFEFEF}\textbf{mBERT} & \cellcolor[HTML]{EFEFEF}\textbf{Mod. mBERT} &  \\
 \cline{2-5}
 \cellcolor[HTML]{EFEFEF} \multirow{-3}{*}{UD En-Hi} & 77.49 & 77.84 & 87.16 & \textbf{88.06} & \multirow{-3}{*}{90.53\textsuperscript{*}}\\
 \hline
  \cellcolor[HTML]{EFEFEF} & \cellcolor[HTML]{EFEFEF}\textbf{Baseline} & \cellcolor[HTML]{EFEFEF}\textbf{Unsup. MUSE} & \cellcolor[HTML]{EFEFEF}\textbf{Sup. MUSE} & \cellcolor[HTML]{EFEFEF}\textbf{BiSkip}  &\cellcolor[HTML]{EFEFEF}\textbf{SOTA}\\
 \cline{2-6}
  \cellcolor[HTML]{EFEFEF} & 60.88 & 60.76 & 60.59 & 60.4 & \\
  \cline{2-5}
 \cellcolor[HTML]{EFEFEF} & \cellcolor[HTML]{EFEFEF}\textbf{BiCVM} & \cellcolor[HTML]{EFEFEF}\textbf{GCM} & \cellcolor[HTML]{EFEFEF}\textbf{mBERT} & \cellcolor[HTML]{EFEFEF}\textbf{Mod. mBERT}& \\
 \cline{2-5}
  \cellcolor[HTML]{EFEFEF} \multirow{-4}{*}{FG En-Hi} & 60.2 & 61.03 & \textbf{63.42} & 63.31 & \multirow{-3}{*}{80.8\footnotemark}  \\
  \hline
  \cellcolor[HTML]{EFEFEF} & \cellcolor[HTML]{EFEFEF}\textbf{Baseline} & \cellcolor[HTML]{EFEFEF}\textbf{Unsup. MUSE} & \cellcolor[HTML]{EFEFEF}\textbf{Sup. MUSE} & \cellcolor[HTML]{EFEFEF}\textbf{BiSkip}  &\cellcolor[HTML]{EFEFEF}\textbf{SOTA}\\
 \cline{2-6}
  \cellcolor[HTML]{EFEFEF} & 88.78 & 88.65 & 88.82 & 89.2 & \\
  \cline{2-5}
 \cellcolor[HTML]{EFEFEF} & \cellcolor[HTML]{EFEFEF}\textbf{BiCVM} & \cellcolor[HTML]{EFEFEF}\textbf{GCM} & \cellcolor[HTML]{EFEFEF}\textbf{mBERT} & \cellcolor[HTML]{EFEFEF}\textbf{Mod. mBERT}& \\
 \cline{2-5}
  \cellcolor[HTML]{EFEFEF} \multirow{-4}{*}{Bangor En-Es} & 87.46 & 89.37 & 93.33 & \textbf{93.62} & \multirow{-3}{*}{95.39\textsuperscript{*}}  \\
\hbl
\end{tabular}}
\caption{\label{POS-results} POS results (F1/\textsuperscript{*}Accuracy) }
\end{table*}
\footnotetext{We create our own test split from the training data, since the test data is not publicly available}
\begin{table*}
\centering
\small
{\renewcommand{\arraystretch}{\pad}
\begin{tabular}{! \vbl l|c|c|c|c|l! \vbl}
\hbl 
 \cellcolor[HTML]{C0C0C0}\textbf{Data} & \cellcolor[HTML]{EFEFEF}\textbf{Baseline} & \cellcolor[HTML]{EFEFEF}\textbf{Unsup. MUSE} & \cellcolor[HTML]{EFEFEF}\textbf{Sup. MUSE} & \cellcolor[HTML]{EFEFEF}\textbf{BiSkip} &\cellcolor[HTML]{EFEFEF}\textbf{SOTA} \\
 \hline
 \cellcolor[HTML]{EFEFEF} & 71.52 & 71.48 & 72.15 & 72.13 & \\
  \cline{2-5}
 \cellcolor[HTML]{EFEFEF}& \cellcolor[HTML]{EFEFEF}\textbf{BiCVM} & \cellcolor[HTML]{EFEFEF}\textbf{GCM} & \cellcolor[HTML]{EFEFEF}\textbf{mBERT} & \cellcolor[HTML]{EFEFEF}\textbf{Mod. mBERT} &  \\
 \cline{2-5}
 \cellcolor[HTML]{EFEFEF} \multirow{-3}{*}{IIITH En-Hi} & 71.55 & 72.37 & 74.96 & \textbf{78.21} & \multirow{-3}{*}{78.14}\\
 \hline
  \cellcolor[HTML]{EFEFEF} & \cellcolor[HTML]{EFEFEF}\textbf{Baseline} & \cellcolor[HTML]{EFEFEF}\textbf{Unsup. MUSE} & \cellcolor[HTML]{EFEFEF}\textbf{Sup. MUSE} & \cellcolor[HTML]{EFEFEF}\textbf{BiSkip}  &\cellcolor[HTML]{EFEFEF}\textbf{SOTA}\\
 \cline{2-6}
  \cellcolor[HTML]{EFEFEF} & 47.9 & 53.74 & 54.17 & 52.98 & \\
  \cline{2-5}
 \cellcolor[HTML]{EFEFEF} & \cellcolor[HTML]{EFEFEF}\textbf{BiCVM} & \cellcolor[HTML]{EFEFEF}\textbf{GCM} & \cellcolor[HTML]{EFEFEF}\textbf{mBERT} & \cellcolor[HTML]{EFEFEF}\textbf{Mod. mBERT}& \\
 \cline{2-5}
  \cellcolor[HTML]{EFEFEF} \multirow{-4}{*}{CALCS En-Es} & 51.6 & 53.57 & 59.69 & \textbf{61.77} & \multirow{-3}{*}{69.17}  \\
\hbl
\end{tabular}}
\caption{\label{NER-results} NER results (F1) }
\end{table*}

\begin{table*}
\centering
\small
{\renewcommand{\arraystretch}{\pad}
\begin{tabular}{! \vbl l|c|c|c|c|l! \vbl}
\hbl 
 \cellcolor[HTML]{C0C0C0}\textbf{Data} & \cellcolor[HTML]{EFEFEF}\textbf{Baseline} & \cellcolor[HTML]{EFEFEF}\textbf{Unsup. MUSE} & \cellcolor[HTML]{EFEFEF}\textbf{Sup. MUSE} & \cellcolor[HTML]{EFEFEF}\textbf{BiSkip} &\cellcolor[HTML]{EFEFEF}\textbf{SOTA} \\
 \hline
 \cellcolor[HTML]{EFEFEF} & 50.44 & 48.37 & 51.27 & 48.84 & \\
  \cline{2-5}
 \cellcolor[HTML]{EFEFEF}& \cellcolor[HTML]{EFEFEF}\textbf{BiCVM} & \cellcolor[HTML]{EFEFEF}\textbf{GCM} & \cellcolor[HTML]{EFEFEF}\textbf{mBERT} & \cellcolor[HTML]{EFEFEF}\textbf{Mod. mBERT} &  \\
 \cline{2-5}
 \cellcolor[HTML]{EFEFEF} \multirow{-3}{*}{SAIL En-Hi} & 49.56 & 50.01 & 58.24 & \textbf{59.35} & \multirow{-3}{*}{56.9}\\
 \hline
  \cellcolor[HTML]{EFEFEF} & \cellcolor[HTML]{EFEFEF}\textbf{Baseline} & \cellcolor[HTML]{EFEFEF}\textbf{Unsup. MUSE} & \cellcolor[HTML]{EFEFEF}\textbf{Sup. MUSE} & \cellcolor[HTML]{EFEFEF}\textbf{BiSkip}  &\cellcolor[HTML]{EFEFEF}\textbf{SOTA}\\
 \cline{2-6}
  \cellcolor[HTML]{EFEFEF} & 50.62 & 58.73 & 58.44 & 60.4 & \\
  \cline{2-5}
 \cellcolor[HTML]{EFEFEF} & \cellcolor[HTML]{EFEFEF}\textbf{BiCVM} & \cellcolor[HTML]{EFEFEF}\textbf{GCM} & \cellcolor[HTML]{EFEFEF}\textbf{mBERT} & \cellcolor[HTML]{EFEFEF}\textbf{Mod. mBERT}& \\
 \cline{2-5}
  \cellcolor[HTML]{EFEFEF} \multirow{-4}{*}{Sentiment En-Es} & 62.62 & 62.89 & 66.03 & \textbf{69.31} & \multirow{-3}{*}{64.6}  \\
\hbl
\end{tabular}}
\caption{\label{Sentiment-results} Sentiment Analysis results (F1) }
\end{table*}

\begin{table*}[!htp]
\centering
\small
{\renewcommand{\arraystretch}{\pad}
\begin{tabular}{! \vbl l|c|c|c|c|l! \vbl}
\hbl 
 \cellcolor[HTML]{C0C0C0}\textbf{Data} & \cellcolor[HTML]{EFEFEF}\textbf{Baseline} & \cellcolor[HTML]{EFEFEF}\textbf{Unsup. MUSE} & \cellcolor[HTML]{EFEFEF}\textbf{Sup. MUSE} & \cellcolor[HTML]{EFEFEF}\textbf{BiSkip} &\cellcolor[HTML]{EFEFEF}\textbf{SOTA} \\
 \hline
 \cellcolor[HTML]{EFEFEF} & 61.39 & 56.11 & 62.78 & 65.56 & \\
  \cline{2-5}
 \cellcolor[HTML]{EFEFEF}& \cellcolor[HTML]{EFEFEF}\textbf{BiCVM} & \cellcolor[HTML]{EFEFEF}\textbf{GCM} & \cellcolor[HTML]{EFEFEF}\textbf{mBERT} & \cellcolor[HTML]{EFEFEF}\textbf{Mod. mBERT} &  \\
 \cline{2-5}
 \cellcolor[HTML]{EFEFEF} \multirow{-3}{*}{QA En-Hi} & 62.33 & 62.78 & \textbf{71.96} & 68.01 & \multirow{-3}{*}{N/A\footnotemark}\\
\hbl
\end{tabular}}
\caption{\label{QA-results} QA results (F1) }
\end{table*}
\footnotetext{The original dataset contains multiple code-mixed pairs and there exists no language based segregation of the results. Since we only choose the EN-HI examples we report this as N/A}
\begin{table*}
\centering
\small
{\renewcommand{\arraystretch}{\pad}
\begin{tabular}{! \vbl l|c|c|l! \vbl}
\hbl 
 \cellcolor[HTML]{C0C0C0}\textbf{Data} & \cellcolor[HTML]{EFEFEF}\textbf{mBERT} & \cellcolor[HTML]{EFEFEF}\textbf{Mod. mBERT} &\cellcolor[HTML]{EFEFEF}\textbf{SOTA} \\
 \hline
 \cellcolor[HTML]{EFEFEF}{NLI En-Hi} & 61.09 & \textbf{63.1} & 63.1 \\
\hbl
\end{tabular}}
\caption{\label{NLI-results} NLI results (Accuracy) }
\end{table*}

\section{Results and Analysis}

Tables \ref{LID-results}-\ref{NLI-results} show the results of using the embedding techniques described above for each task and dataset. mBERT provides a large increase in accuracy as compared to cross-lingual techniques, and in most cases, the modified mBERT technique performs best. We do not experiment with baseline or cross-lingual embedding techniques for NLI, since we find that mBERT surpasses the other techniques for all other tasks. For NLI, as in the other cases, we find that modified mBERT performs better than mBERT. We hypothesize that this happens because code-switched languages are not just a union of two monolingual languages. The distributions and usage of words in code-switched languages differ from their monolingual counterparts, and can only be captured with real code-switched data, or synthetically generated data that closely mimics real data.

\cite{glavas2019properly} point out how all cross-lingual word embedding methods optimize for bilingual lexicon induction. Each model is trained using different language pairs and different training and evaluation dictionaries, leading to it overfitting to the task it is optimizing for and failing in other cross-lingual scenarios. Also, the loss function in training cross-lingual word embeddings has a component where $w1$ in one language predicts the context of its aligned word $w2$ in the other language. However, in the case of code-switching, $w1$ appearing in the context of $w2$ may not be natural. This clearly highlights the need to learn cross-lingual embeddings keeping code-mixed language processing as an optimization objective.

%In addition, in the process of projecting the embeddings to a common space, a word from one language is made to predict the context of its equivalent word in the other language which is not necessarily how code-switched languages behave. This clearly highlights the need to learn cross-lingual embeddings keeping code-mixed language processing as an optimization objective.

The results using mBERT cannot be directly compared to the cross-lingual models because of the difference in the magnitude of data involved in training. Also, due to the fact that mBERT is trained on 104 languages together, with massive amounts of data for a large number of epochs, it learns several common features better providing for a well represented common embedding space. The training data used for training the cross-lingual embeddings is restricted to Twitter and query logs, while mBERT is trained on the entire wiki dump.

% Our domain was also constricted (twitter for en-es and bing for en-hi) while mbert encompasses a more generic domain given it is entire wikidump.

Overall, the cross-lingual and mBERT models perform better for \textit{English-Spanish} as compared to \textit{English-Hindi}. This could be due to several reasons. 
\begin{itemize}
    \item English and Spanish are similar languages, with both mostly retaining individual word order while code-switching, which is not the case for English and Hindi.
    \item Romanized Hindi does not use standardized spellings, and errors made by the transliterator could have influenced the results.
    \item We use Twitter and social media data to train cross lingual word embeddings for \textit{English-Spanish} which are similar in domain to the task datasets, while we use the IITB and query-based parallel corpora for \textit{English-Hindi} which is generic in domain, constrained by the available resources at hand. 
\end{itemize}

We find that for most tasks, modified mBERT performs better than mBERT. In cases where this is not true (QA En-Hi; FG En-Hi), the difference in accuracy between the two models is small. This could be attributed to errors made by the transliterator or corpus differences, but in general we observe that the \textit{modified En-Hi} mBERT model does not significantly outperform the base mBERT model. Given the promising results obtained by modified mBERT, it would be interesting to pre-train a language model for code-switched data which is trained on the monolingual corpora of languages involved and fine-tuned on GCM as proposed, to compare against fine-tuning mBERT itself, which is trained on multiple languages.

We find that accuracies vary across tasks in the GLUECoS \, benchmark, and except in the case of LID, code-switched NLP is far from solved. This is particularly stark in the case of Sentiment and NLI, which are three and two way classification tasks respectively. Modified mBERT performs only a little over chance, which shows that we are still in the early days of solving NLI for code-switched languages, and also indicates that our models are far from truly being able to understand code-switched language. 
% We plan to release the NLI dataset to the community, which we hope will spur research in this area. 

\section{Conclusion}

In this paper, we introduce the first evaluation benchmark for code-switching, GLUECoS. The benchmark contains datasets in \textit{English-Hindi} and \textit{English-Spanish} for six NLP tasks - LID, POS tagging, NER, Sentiment Analysis, Question Answering and a new code-switched \textit{Natural Language Inference} task. We test various embedding techniques across all tasks and datasets and find that multilingual BERT outperforms cross-lingual embedding techniques on all tasks. We also find that for most datasets, a modified version of mBERT that has been fine-tuned on synthetically generated code-switched data with a small amount of real code-switched data performs best. This indicates that while multilingual models do go a long way in solving code-switched NLP, they can be improved further by using real and synthetic code-switched data, since the distributions in code-switched languages differ from the two languages being mixed.

In this work, we use standard architectures to solve each NLP task individually and vary the embeddings used. In future work, we would like to experiment with a multi-task setup wherein tasks with less training data can significantly benefit from those having abundant labelled data, since most code-switched datasets are often small and difficult to annotate. We experiment with datasets having varied amounts of code-switching and from different domains and show that some tasks, such as LID and POS tagging are relatively easier to solve, while tasks such as QA and NLI have low accuracies. We would like to add more diverse tasks and language pairs to the GLUECoS benchmark in a future version.

All the datasets used in the GLUECoS benchmark are publicly available, and we plan to make the NLI dataset available for research use. We hope that this will encourage researchers to test multilingual, cross-lingual and code-switched embedding techniques and models on this benchmark.
\bibliography{anthology,acl2020}
\bibliographystyle{acl_natbib}
\newpage
\appendix

\label{sec:appendix}
\section{Additional Dataset Details}
\label{app:dataset}
For each dataset wherein the training, development and test splits are not provided, we create balanced custom splits in an 8:1:1 ratio.

For En-Hi datasets, where the corpus is originally in the Roman script and is not language tagged, we use the LID tool provided by \cite{rijhwani2017estimating} to obtain language tags.

In cases where language tags are provided, we convert Roman Hindi words to Devanagari using an off-the-shelf transliterator.

\section{Additional Training Details}
\label{app:train}
We conduct each experiment for 5 random seed values and report the average of the results obtained.
\subsection{Word-Level Tasks}

For the word level tasks including Language Identification, Named Entity Recognition and Part of Speech tagging we make use of the sequence labeler.\footnote{https://github.com/marekrei/sequence-labeler/tree/484a6beb1e2a2cccaac74ce717b1ee30c79fc8d8} This implements a BiLSTM with a CRF layer on the top as described in \cite{lample2016neural}. We use the adadelta optimizer with a learning rate of 1.0, dropout of 0.5 and a batch size of 32. We run the model for a maximum of 20 epochs and stop if the validation accuracy on the \textit{best\_model\_selector} hyperparameter shows no improvement for 5 epochs continually. The \textit{best\_model\_selector} hyperparameter is the F1 score. The dimension of the word embeddings is 300.

We make use of the \textit{transformers} library\footnote{https://github.com/huggingface/transformers} for the mBERT experiments. We use the AdamW optimizer with a learning rate of 5e-5, epsilon of 1e-8, and a batch size of 32, as suggested by \cite{devlin2018bert}. We train for 5 epochs.

\subsection{Sentence-Level Tasks}
For the sentence level tasks (Sentiment Classification) we implement a BiLSTM with one hidden layer of dimension 256. We apply a dropout of 0.5,  and use the Adam optimizer with a 0.001 learning rate and 1e-8 epsilon value. We use a batch size of 64 and train for a maximum of 15 epochs stopping if the validation accuracy continually drops for 3 epochs. The dimension of the word embeddings is 300.

We make use of the \textit{transformers} library for the mBERT experiments. We use the AdamW optimizer with a learning rate of 5e-5, epsilon of 1e-8, and a batch size of 32, as suggested by \cite{devlin2018bert}. We train for 5 epochs.

\subsection{Sentence-Pair Tasks}
For the embedding evaluations on the QA task, we make use of the BiDAF architecture\footnote{https://github.com/ElizaLo/Question-Answering-based-on-SQuAD} as proposed in \cite{seo2016bidirectional}. We keep the default training hyperparameters which include a learning rate of 0.5, a batch size of 1, training epochs as 5, a maximum context length of 400 tokens and a maximum question length of 50 tokens.

We make use of the SQuAD training script and the XNLI training script of the Transformers library with its default hyperparameters for the mBERT experiments.

\section{BERT LM fine-tuning}
\label{app:bert}
We take the bert model released by Google (\textit{bert-base-multilingual-cased}) and fine-tune it for masked language modeling on 2 types of code-mixed datasets. 

We use a curriculum wherein the model is first trained on generated code-mixed data (gCM) for 10 epochs and then on real code-mixed data (rCM) for 10 epochs.

For English-Hindi, the details of the datasets are as follows:
\begin{itemize}
    \item 2M gCM sentences generated from the parallel corpus by \newcite{kunchukuttan2018iit}
    \item 93k rCM sentences from the corpora by \newcite{chandu-etal-2018-language}
\end{itemize}

For English-Spanish, the details of the datasets are as follows:
\begin{itemize}
    \item 8M gCM sentences generated from the corpus by \newcite{rijhwani2017estimating}
    \item 93k rCM sentences from the corpus by \newcite{rijhwani2017estimating}
\end{itemize}
\end{document}